\documentclass{article}
\usepackage{ijcai24}

\usepackage{times}
\usepackage{soul}
\usepackage{url}
\usepackage[hidelinks]{hyperref}
\usepackage[utf8]{inputenc}
\usepackage[small]{caption}
\usepackage{graphicx}
\usepackage{amsmath}
\usepackage{amsthm}
\usepackage{booktabs}
\usepackage{algorithm}
\usepackage{algorithmic}
\usepackage[switch]{lineno}
\usepackage{multirow}
\usepackage{multicol}
\usepackage{subfigure}
\usepackage[table]{xcolor}

\title{CLIP-Driven Outliers Synthesis for Few-Shot Out-of-Distribution Detection}

\author{
Hao Sun$^1$
\and
Rundong He$^{1*}$\and
Zhongyi Han$^{2}$\and
Zhicong Lin$^1$\and
Yongshun Gong$^1$\and
Yilong Yin$^{1*}$\\
\affiliations
$^1$Shandong University\\
$^2$King Abdullah University of Science and Technology\\
\emails
sunhao\_@mail.sdu.edu.cn
}

\begin{document}

\maketitle

\begin{abstract}

    Few-shot OOD detection focuses on recognizing out-of-distribution (OOD) images that belong to classes unseen during training, with the use of only a small number of labeled in-distribution (ID) images. Up to now, a mainstream strategy is based on large-scale vision-language models, such as CLIP. However, these methods overlook a crucial issue: the lack of reliable OOD supervision information, which can lead to biased boundaries between in-distribution (ID) and OOD. To tackle this problem, we propose CLIP-driven Outliers Synthesis~(CLIP-OS). Firstly, CLIP-OS enhances patch-level features' perception by newly proposed patch uniform convolution, and adaptively obtains the proportion of ID-relevant information by employing CLIP-surgery-discrepancy, thus achieving separation between ID-relevant and ID-irrelevant. Next, CLIP-OS synthesizes reliable OOD data by mixing up ID-relevant features from different classes to provide OOD supervision information. Afterward, CLIP-OS leverages synthetic OOD samples by unknown-aware prompt learning to enhance the separability of ID and OOD. Extensive experiments across multiple benchmarks demonstrate that CLIP-OS achieves superior few-shot OOD detection capability.
\end{abstract}

\section{Introduction}


Detecting out-of-distribution (OOD) samples is essential for the secure deployment of machine learning models in real-world environments, as the occurrence of new categories requiring special treatment is inevitable. Traditional OOD detection methods rely on single-model supervised learning~\cite{hendrycks2016baseline,hendrycks2019scaling,he2022ronf,he2023topological}, which has yielded promising results. In the financial sector, OOD detection can assist in identifying unseen transaction patterns, thereby reducing the risk of fraud. In medical imaging analysis, OOD detection can help identify rare cases that deviate from common medical conditions.

However, the current focus of OOD detection is mainly on fully supervised or zero-shot methods. Fully supervised approaches~\cite{tao2023non} require large-scale in-distribution~(ID) training samples as well as training resources, which are often difficult to obtain in practice. On the other hand, zero-shot OOD detection~\cite{fort2021exploring,esmaeilpour2022zero} lacks specific task supervision and struggles to address potential domain gaps, consequently limiting its effectiveness in OOD detection. In response to these challenges, few-shot OOD detection emerges as a critical solution, utilizing only a small subset of ID training samples for learning~\cite{jeong2020ood}.

\begin{figure}[t] 
\setlength{\abovecaptionskip}{0.1cm}
    \centering
    \includegraphics[width=0.46\textwidth]{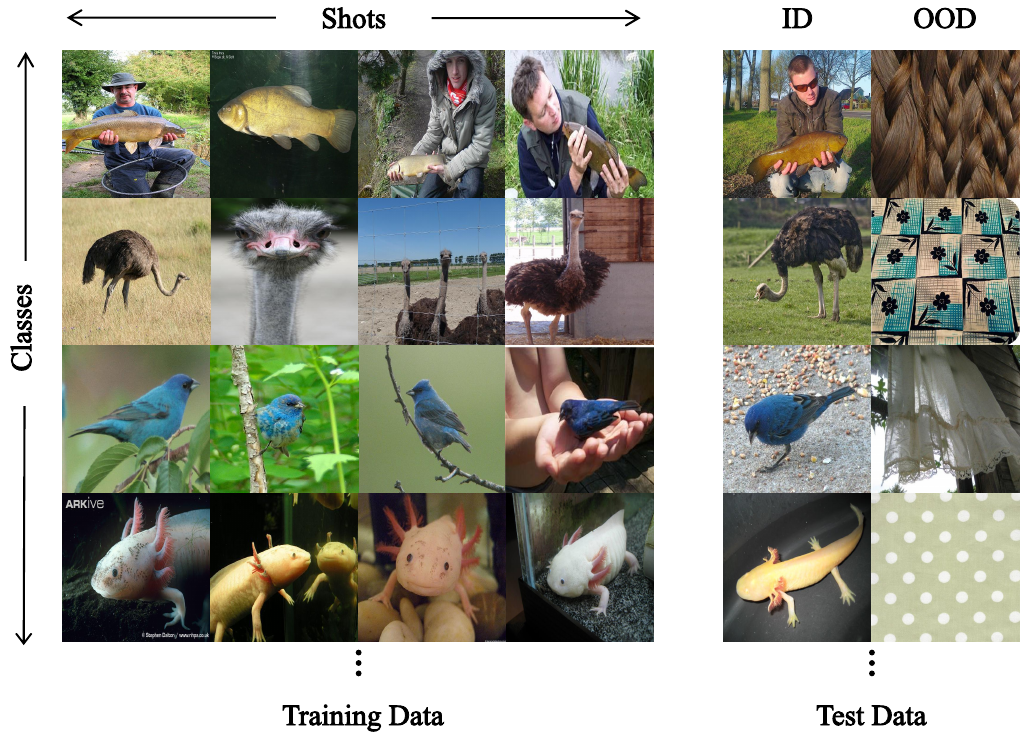} 
    \caption{Few-shot OOD detection. We show some training data of ImageNet in four-shot setting and 8 test data from the test set of ImageNet and Textures. The goal is to correctly classify the ID samples while also detecting the OOD samples.}
    

    \label{fig:setting} 
\end{figure}
Currently, there are two main approaches to address few-shot OOD detection: traditional methods and those based on large vision-language models. Traditional methods~\cite{jeong2020ood} essentially rely on meta-learning, using a small number of samples to train models that can effectively generalize when faced with unknown categories. With the development of large models, their feature extraction capabilities have been significantly enhanced, drawing our attention to these large models. The current mainstream methods based on large models include LoCoOp~\cite{miyai2023locoop} and CLIPood\cite{shu2023clipood}, both of which involve fine-tuning CLIP to address downstream tasks. LoCoOp, specifically designed for few-shot OOD, generates synthetic OOD data using only background information, primarily enhancing the recognition of background as OOD while facing challenges in learning the boundaries between real IDs categories and OOD categories. CLIPood utilizes fine-tuning of CLIP to address downstream classification tasks and has the potential to solve few-shot OOD problems, but it does not incorporate synthetic OOD data for model training. These methods lack effective OOD supervision to guide the learning of unbiased boundaries between ID and OOD. Therefore, the fundamental problem is \textbf{how to synthesize reliable OOD supervised signals with large model under few-shot scenario}.




Based on the above motivation, we propose CLIP-driven Outliers Synthesis~(CLIP-OS) to synthesize reliable OOD supervised signals for few-shot OOD detection. Since ID-irrelevant features of the ID data (such as background) lead to noisy ID feature distribution, thereby hindering the learning of ID-class boundaries. Therefore, CLIP-OS firstly aims to obtain ID-relevant features by patch-context incorporating, which enhances the perception of features around patches, and CLIP-surgery-discrepancy masking, which separate ID-relevant and ID-irrelevant features adaptively. Then, CLIP-OS synthesizes reliable OOD data by mixing up ID-relevant features from different ID classes. Finally, CLIP-OS utilizes synthetic OOD data for regularizing the boundary between ID and OOD by unknown-aware prompt learning, which aligns the OOD supervised signals with the text
embedding of the ``unknown'' prompt.



Through experiments on CIFAR-10, CIFAR-100, and ImageNet, we observe that our method significantly outperforms existing approaches. Even with very few samples, such as one-shot scenario, our method achieves substantial progress, even surpassing fully supervised methods. Our contributions can be summarized as follows: 

\begin{itemize}
    \item We delve into an unexplored problem: how to synthesize reliable OOD supervised signals with CLIP under few-shot scenario. Based on this, we propose CLIP-OS for few-shot OOD detection.

    \item We propose patch-context learning to enhance the perception of features around patches and CLIP-surgery-discrepancy masking to separate ID-relevant and ID-irrelevant features adaptively.

    \item We synthesize reliable OOD data by mixing up ID-relevant features from different ID classes and utilize them for regularizing the boundary between ID and OOD by unknown-aware prompt learning.

    \item Extensive experiments demonstrate that CLIP-OS achieves remarkable improvements compared to previous approaches, which also verifies the effectiveness of synthesizing reliable OOD supervised signals.




\end{itemize}

\begin{figure*}[ht]

\centering
\includegraphics[height=7.5cm,width=17cm]{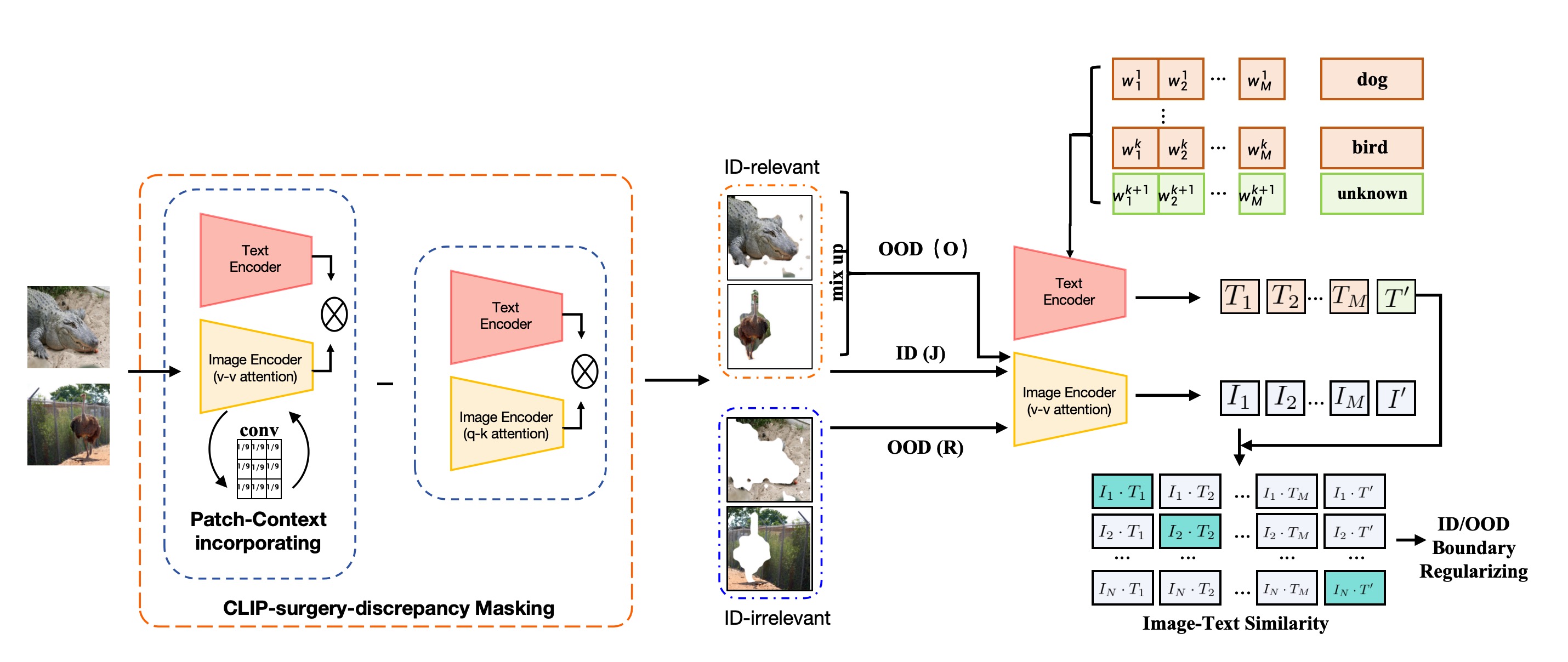}
\caption{The overall framework of our proposed CLIP-driven Outliers Synthesis.}

\label{fig:framework}
\end{figure*}

\section{Related Work}

\subsection{OOD detection}

The distinction between in-distribution~(ID) and out-of-distribution~(OOD) data is crucial for the deployment of machine learning models in real-world applications. OOD detection methods can be mainly categorized into two types: classification-based methods~\cite{DBLP:conf/nips/LeeLLS18,DBLP:conf/nips/LiuWOL20,DBLP:conf/icml/SunM0L22,DBLP:conf/icml/WeiXCF0L22,DBLP:conf/iclr/DuWCL22}  and density-based methods~\cite{DBLP:conf/nips/XiaoYA20,DBLP:conf/aistats/MorningstarHGLA21,DBLP:conf/icml/ZhouL21,DBLP:conf/iclr/JiangSY22,DBLP:conf/icml/ZhangGR21}. Classification-based methods model the conditional distribution of ID data and design a scoring function to measure the uncertainty of test data, while density-based methods use probabilistic models to represent the ID distribution and identify test data in low-density regions as OOD. However, density-based OOD detection methods are often challenging to train and optimize, and they generally exhibit inferior performance compared to classification-based methods~\cite{DBLP:journals/corr/abs-2110-11334}. Therefore, in this paper, we focus on classification-based methods. This category encompasses two primary research branches: testing-time methods~\cite{DBLP:conf/nips/LeeLLS18,DBLP:conf/nips/LiuWOL20,DBLP:conf/cvpr/Wang0F022,DBLP:conf/nips/ZhuCXLZ00ZC22,DBLP:journals/corr/abs-2209-08590,sun2022out} and training-time methods~\cite{DBLP:conf/iclr/HendrycksMD19,DBLP:conf/icml/WeiXCF0L22,DBLP:conf/iclr/DuWCL22,DBLP:conf/cvpr/HeHLY22,DBLP:conf/nips/WangLZZ0L022}. Testing-time methods are convenient to use without modifying the training procedure and objective~\cite{DBLP:journals/corr/abs-2110-11334}. In contrast, training-time methods aim to alleviate overconfident predictions for OOD data during training. Based on whether OOD-supervised signals are utilized in the training process, training-time methods can be classified into OOD-free and OOD-needed methods. Examples of OOD-free methods include those proposed by Wei et al.~\cite{DBLP:conf/icml/WeiXCF0L22}, where LogitNorm was incorporated into the cross-entropy loss to isolate the influence of logits’ norm from the training process, and by Lin et al.~\cite{DBLP:conf/cvpr/LinRL21}, who leveraged intermediate classifier outputs for dynamic and efficient OOD inference. On the other hand, OOD-needed methods seek to calibrate the model by using OOD-supervised signals obtained from auxiliary OOD datasets~\cite{DBLP:conf/iclr/HendrycksMD19,DBLP:conf/nips/LiuWOL20,DBLP:conf/pkdd/ChenLWLJ21,DBLP:conf/icml/MingFL22}, unlabeled data~\cite{DBLP:conf/cvpr/HeHLY22,DBLP:conf/iccv/YuA19,DBLP:conf/iccv/YangWFYZZ021,DBLP:conf/nips/ZhouGCLP21,DBLP:conf/icml/Katz-SamuelsNNL22}, or synthetic virtual OOD data~\cite{DBLP:conf/iclr/DuWCL22,DBLP:conf/iccv/TangMPWSGTW21,DBLP:conf/nips/TackMJS20,DBLP:conf/mm/HeHLY22}. It is important to note that previous assumptions were based on a large sample size, which may be challenging to meet in practice, making it necessary to explore few-shot learning.

\subsection{Few-shot OOD detection}

To address the challenge of obtaining large amounts of data and computational resources, which are often difficult to achieve in traditional OOD detection, few-shot OOD detection has been proposed. Traditional methods rely on a single model and meta-learning to obtain boundaries between in-distribution (ID) and out-of-distribution (OOD) data~\cite{jeong2020ood,wang2020few}.FS-OOD~\cite{wang2020few} proposed the Out-of-Episode Classifier~(OEC) and leveraged examples from the same dataset but not from the current episode to achieve few-shot OOD detection. OOD-MAML leveraged model-agnostic meta-learning~(MAML) and synthesized OOD examples through gradient updates of specific meta-parameters to assist in learning a better boundary.  
However, with the development of large-scale vision-language models, there is increasing interest in leveraging these models for few-shot OOD detection. 
Currently, the mainstream approaches are based on large models and utilize fine-tuning on a small number of samples to achieve OOD detection. LoCoOp~\cite{miyai2023locoop}, building upon prompt learning, used selected patches as OOD samples for OOD supervision. CLIPood~\cite{shu2023clipood} employed fine-tuning techniques and trained the model using proposed Margin Metric Softmax and Beta Moving Average. However, they both suffer from a common issue - the lack of effective OOD supervision guidance. CLIPood does not synthesize OOD examples to guide training, while LoCoOp, despite synthesizing OOD samples, uses background information that does not effectively enhance class discriminability.




\section{Method}

\subsection{Problem statement}

In the OOD detection setting, the term ``ID classes" refers to the classes present in the training data. Correspondingly, the OOD classes are those that do not belong to any of the ID classes, indicating that the model has not been exposed to them during the training phase. Formally, we assume an ID dataset $D^{in}$ consisting of $(x^{in}, y^{in})$ samples, where $x^{in}$ represents input ID images and $y^{in}\in Y^{in}:=1,...,M$ represents ID class labels, $M$ denotes number of ID classes. Similarly, the OOD dataset $D^{in}$ consists of $(x^{out}, y^{out})$ samples, with $x^{out}$ representing input OOD images and $y^{out}\in Y^{out}:=M+1,...,M+N$ representing OOD class labels, $N$ denotes number of OOD classes. It is important to note that there is no overlap between $Y^{in}$ and $Y^{out}$, meaning their intersection is an empty set, $Y^\mathrm{in}\cap Y^\mathrm{out}=\emptyset$.

The scenario under investigation involves fine-tuning vision-language models on the few-shot training set without exposure to any OOD data. The test set used to evaluate the model's OOD detection capability includes both $D^{in}$ and $D^{in}$. In contrast to traditional zero-shot methods and fully supervised methods using all training data, we only selected a small number of samples for each class in $D^{in}$ during training (one and two samples per class).

\subsection{CLIP-driven Outliers Synthesis}

Since the existing few-shot OOD detection methods lack effective OOD supervision, resulting in biased boundaries between ID and OOD. Therefore, the fundamental problem is how to synthesize reliable OOD supervised signals under few-shot scenario with the help of CLIP model. Based on this, we propose CLIP-driven Outliers Synthesis (CLIP-OS) to synthesize reliable OOD supervised signals for few-shot OOD detection. An overview of CLIP-OS is illustrated in Fig.~\ref{fig:framework}, which mainly contain three steps: (1) obtaining ID-relevant features; (2) synthesizing reliable OOD data; (3) regularizing ID/OOD boundary.

\subsubsection{ID-relevant Features Obtaining}
Since ID-irrelevant features of the ID data (such as background) lead to noisy ID feature distribution, thereby hindering the learning of ID-class boundaries. Therefore, CLIP-OS obtain ID-relevant features by patch-context incorporating, which enhances the perception of features around patches, and CLIP-surgery-discrepancy masking, which separate ID-relevant and ID-irrelevant features adaptively.

\paragraph{Patch-Context Incorporating.}



The existing vision-language models, when separating the ID-relevant and ID-relevant areas based on patches, need to calculate the similarity between patch embeddings and class text embeddings. However, current methods only focus on the information of a single patch, resulting in overly localized patch embeddings. The information contained in each independent patch lacks discriminative power, leading to classification errors for some patches. To address this issue, we incorporate the patch-context's embedding into the current patch's embedding.

However, there is still a problem about computational efficiency. If a loop is used to iterate through the patches, the update of a large number of patches in the network will make the computation time unbearable. To address this, we designed a fixed 3×3 convolution kernel, where each element is 1/9, to utilize GPU matrix operations to accelerate this process. For each patch in the Attention module, we pass it through a convolution operation to incorporate the average information of the surrounding eight patches, ensuring that each patch includes information from its surroundings. Specifically, the patch-context incorporating is as follows:
\begin{equation}
    v_{ij} = v_{ij} + \beta\textit{conv}(v_{ij})\,,
\end{equation}
where $v_{ij}$ represents the value of the patch at the $i$-th row and $j$-th column, $\beta$ denotes the hyperparameter and $\textit{conv}$ denotes the convolution operation.

\paragraph{CLIP-surgery-discrepancy Masking.}

To achieve the separation of ID-relevant and ID-irrelevant regions, LoCoOp initially relies on a pre-trained CLIP to calculate the similarity between all patch embeddings and all class text embeddings. If the similarity between the current patch and the real label's text embedding ranks in the top $K$ among all class text embeddings, then the current patch is considered an ID-relevant region. However, this strategy has two critical issues. First, the hyperparameter $K$ is very sensitive and difficult to set optimally. Using a fixed $K$ for each sample leads to a suboptimal selection of ID-relevant and ID-irrelevant regions, as a fixed $K$ struggles to handle the complex and varied nature of samples, such as those with fewer foreground areas or multiple foreground areas. Therefore, exploring an adaptive strategy for separating ID-relevant and ID-irrelevant regions is crucial. Second, according to Li et al.~\shortcite{li2023clip}, CLIP leads to an opposite visualization. The opposite visualization refers to the visualization results that are opposite to the ground-truth. For example, the similarity between the embeddings of background regions and the corresponding real label's text embedding is greater than the similarity between the embeddings of foreground regions and the corresponding real label's text embedding. To address these two issues, we propose the CLIP-surgery-discrepancy Masking.

Li et al.~\shortcite{li2023clip} discovers that the major cause of opposite visualization is the parameters (query and key) in the self-attention module, which heavily focus on opposite semantic regions. To address this issue, CLIP-surgery-discrepancy Masking replaces the original self-attention via the proposed v-v self-attention, which is defined by
\begin{equation}
    Attn_{vv} = \textit{softmax}(V \cdot V^T \cdot scale) \cdot V .
\end{equation}

We utilize vision-to-vision (v-v) self-attention to obtain the embedding of each patch, and then calculate the similarity between each patch's embedding and the ground-truth class text embedding to generate a similarity map $SMap$. Based on the similarity map, we can obtain the ID-relevant region $J$ and the ID-irrelevant region $R$ by
\begin{equation}
\begin{aligned}
    &J = \{P_{ij} \in I : SMap(i,j)>0.5\} \,,\\
    &R = \{P_{ij} \in I : (i,j)\notin J\}\,,
    \label{JR}
    \end{aligned}
\end{equation}
where $P_{ij}$ denotes the patch in the $i$-th row and $j$-th column, $I$ denotes the original input, $SMap(i,j)$ denotes represents the similarity score at the $i$-th row and $j$-th column in the similarity map $SMap$. 

Although we can obtain the ID-relevant region $J$ and the ID-irrelevant region $R$ based on Eq.~\eqref{JR} effectively, we cannot handle the boundaries between ID-relevant and ID-irrelevant regions well, due to the similarity scores being relatively close at these boundary regions.  To address this issue, we skillfully utilize the characteristics of CLIP's opposite visualization, meaning that patches more representative of the foreground tend to have lower similarity scores. By calculating the difference between the our $SMap$ and CLIP's $SMap_{clip}$, we can enhance the score of the foreground at the boundaries and reduce it for the background, thereby obtaining a more accurate boundary. We use CLIP to obtain the opposite similarity map:
\begin{equation}
 SMap_{clip} = f_{img} @ f_{text} \,,
\end{equation}
where ``@" represents matrix multiplication, and $f_{img}$ and $f_{text}$ respectively represent the features of image patches and text. We then incorporate the difference between the two to increase the distinction between the foreground and background at the boundaries by 
\begin{equation}
\begin{aligned}
    J = &\{P_{ij} \in I : SMap(i,j) \\
    & + (SMap(i,j) - SMap_{clip}(i,j)) * 0.1)  >0.5\} \,.
\label{JR2}
\end{aligned}
\end{equation}

\subsubsection{Reliable OOD Data Synthesizing}

Previous research underscores the necessity of Out-of-Distribution (OOD) supervision signals for effective learning. Many studies employ large-scale auxiliary OOD datasets as such signals. Yet, acquiring these extensive auxiliary OOD datasets in real-world applications poses significant challenges, including increased computational and storage demands, and the potential introduction of noise. Moreover, a distribution shift often occurs between OOD data encountered during testing and the OOD auxiliary dataset used in training, leading to less than optimal OOD detection performance. Consequently, generating OOD data from in-distribution (ID) data emerges as a viable approach. Traditional methods for synthesizing OOD data from ID data typically involve modeling the Gaussian distribution of ID features using extensive ID data. However, these methods are not feasible in few-shot scenarios where modeling a Gaussian distribution with limited data (few-shot or one-shot settings) is impractical.

To facilitate effective OOD data synthesis in few-shot scenarios, this study leverages ID-relevant features identified in the prior step and implements a mixup of cross-class ID-relevant features by 

\begin{equation}
\label{eqs:oodid}
    O=\{o\mid o=\lambda\mathbf{v}_i+(1-\lambda)\mathbf{v}_j,i\neq j\}\,,
\end{equation}
where $\mathbf{v}$ is the feature of $J$, and $i,j$ represent different ID classes respectively. With Eq.~\eqref{eqs:oodid} we obtain reliable OOD samples for OOD-supervised training.



\subsubsection{ID/OOD Boundary Regularizing}

The absence of OOD supervised signals can lead to biased boundaries, resulting in an underestimation of uncertainty for OOD samples and potentially causing OOD samples to be misclassified as ID classes. Therefore, regularing ID/OOD boundary with OOD supervised signals is necessary. The OOD loss in LoCoOp is achieved by maximizing the entropy of OOD samples, that is, evenly distributing the predicted probabilities to all seen classes. The uniform distribution is unsuitable to calibrate ID/OOD Boundary. For example, the unseen class ``cat'' is more similar to the dog than the bird, thus, this learning scheme with the same constraint for all unseen-class is not reasonable. Since OOD detection and ID classification are highly interrelated, the uniform distribution regularizing impairs the performance of ID classification, which in turn affects the performance of OOD detection.

 
To address this issue, we propose unknown-aware prompt learning for regularizing the boundary between ID and OOD, which aligns the OOD supervised signals with the text embedding of the ``unknown" prompt. In this way, we ensure both the model's classification performance on ID classes and improve the OOD detection capability, as OOD samples are classified as ``unknown" without affecting the training results of ID category prompt embeddings. we use cross-entropy loss to calculate OOD loss as our OOD regularization: 

\begin{equation}
    \label{eqs:crossentropy1}
    \mathcal{L}_{ood} = - \log \frac{\exp\left(\textit{sim}\left(\boldsymbol{f}_o,\boldsymbol{g}_{M+1}\right)/\tau\right)}{\sum_{m=1}^{M+1}\exp\left(\textit{sim}\left(\boldsymbol{f}_o,\boldsymbol{g}_{m}\right)/\tau\right)}\,,
\end{equation}
where $\boldsymbol{f}_o$ denotes the feature of OOD data $\boldsymbol{x}_o$, $\boldsymbol{x}_o \in O \cup R$, $\boldsymbol{g}_{M+1}$ denotes the text embedding of ``unknown" prompt. Similarly, ID loss can be obtained by 
\begin{equation}
    \label{eqs:crossentropy2}
    \mathcal{L}_{id} = - \log \frac{\exp\left(\textit{sim}\left(\boldsymbol{f}_i,\boldsymbol{g}_{y}\right)/\tau\right)}{\sum_{m=1}^{M+1}\exp\left(\textit{sim}\left(\boldsymbol{f}_i,\boldsymbol{g}_{m}\right)/\tau\right)}\,,
\end{equation}
where $\boldsymbol{f}_i$ denotes the feature of ID data $\boldsymbol{x}_i$, $\boldsymbol{x}_i \in J$.

Then, by combining Eqs.~\eqref{eqs:crossentropy1} and \eqref{eqs:crossentropy2}, we arrive at the final optimization objective as follows,
\begin{equation}
    \label{eqs:final}
    \mathcal{L} = \mathcal{L}_{id} + \beta\mathcal{L}_{ood},
\end{equation}
where $\beta$ is a hyperparameter.

\begin{table*}[t]
\setlength{\abovecaptionskip}{0.1cm}
\centering
\scalebox{0.87}{
\begin{tabular}{l|cccccc|cccccc}
\toprule 
ID data & \multicolumn{6}{c|}{CIFAR-10} & \multicolumn{6}{c}{CIFAR-100}\\
\midrule 
OOD data & Textures  & Places  & LSUN-C  &LSUN-R &iSUN  & Avg  & Textures  & Places  & LSUN-C  &LSUN-R &iSUN  & Avg  \\
\midrule

&\multicolumn{12}{c}{\textit{zero-shot}} \\
MCM  &92.25   &80.41  &92.72 &82.73  &82.08 &86.03  &60.15 &55.16  &85.99  &82.25 &77.82 &72.27 \\
GL-MCM &92.89 &82.48 &90.21 &82.02 &80.04 &85.53 &70.76 &61.90 &82.03 &74.23 &69.03 &71.59 \\

\midrule
&\multicolumn{12}{c}{\textit{one-shot}} \\
CLIPood &67.94 &14.16 &89.53 &87.50 &89.23 &69.67 &\textbf{81.46} &25.89 &89.04 &81.81 &83.61 & 72.36\\
CoOp  &87.69 &72.55 &97.33 &97.36 &95.95 &90.18 &62.29 &49.45 &87.03 &82.72 &81.54 &72.60  \\
LoCoOp &89.65 &84.76 &96.68 &98.02 &97.20 &93.26 &64.08 &54.20 &\textbf{90.22} &83.11 &82.24 &74.77 \\
\rowcolor{black!20} CLIP-OS (ours) &\textbf{93.63} &\textbf{87.65} &\textbf{98.17} &\textbf{98.46} &\textbf{97.91} &\textbf{95.16} &70.95 &\textbf{59.04} &89.60 &\textbf{85.71} &\textbf{85.88} &\textbf{78.24}\\

\midrule
&\multicolumn{12}{c}{\textit{two-shot}} \\
CLIPood &64.66 &11.65 &91.25 &90.64 &90.44 &69.73 &\textbf{82.19} &28.16 &85.73 &80.40 &82.80 & 71.84\\
CoOp &83.41 &85.54 &97.49 &97.82 &97.13 &92.28 &61.64 &47.47 &84.07 &79.89 &80.61 &70.74   \\
LoCoOp &90.52  &84.00 &\textbf{98.19} &98.33 &97.76 &93.76 &62.43 &57.55 &86.44 &81.43 &82.60 &74.09 \\
\rowcolor{black!20} CLIP-OS (ours) &\textbf{93.50} &\textbf{88.04} &98.12 &\textbf{98.51} &\textbf{98.00} &\textbf{95.23} &70.88 &\textbf{60.04} &\textbf{87.50} &\textbf{85.69} &\textbf{85.82} &\textbf{77.99}\\

\bottomrule
\end{tabular}}
\caption{
Comparison results on CIFAR-10 and CIFAR-100 OOD benchmarks. 
}
\label{table:main}
\end{table*}

\begin{table}[t]
\setlength{\abovecaptionskip}{0.1cm}
\centering
\scalebox{0.75}{
\begin{tabular}{l|cccccc} 
\toprule 
ID Data & \multicolumn{6}{c}{ImageNet-100}\\
\midrule
OOD data   & Textures    & Places   & LSUN-C    & LSUN-R    & iSUN   & Avg  \\
\midrule
&\multicolumn{6}{c}{\textit{zero-shot}} \\
MCM     &71.03   &70.25   &96.44   &87.42    &89.64    &83.59  \\
GL-MCM  &71.03  &69.24  &89.59 &74.29  &79.49  &76.73 \\
\midrule
&\multicolumn{6}{c}{\textit{one-shot}} \\
LoCoOp   &\textbf{76.26} &66.38 &96.86  &91.68  &94.88  &85.21 \\
\rowcolor{black!20} CLIP-OS (ours)  &71.11  &\textbf{69.82}  &\textbf{97.51}  &\textbf{97.68}  &\textbf{97.54} &\textbf{86.73}\\
\bottomrule
\end{tabular}}
\caption{
Comparison results on ImageNet-100. 
}
\label{table:imagenet}
\end{table}

\section{Experiments}

\subsection{Experimental Detail}

\paragraph{Datasets.}We use CIFAR-10\cite{krizhevsky2009learning}, CIFAR-100\cite{krizhevsky2009learning}, and ImageNet-100 as the ID data. ImageNet-100 consists of 100 randomly sampled classes from the ImageNet\cite{deng2009imagenet} dataset. For the OOD dataset, we used 5 common OOD datasets: Textures\cite{cimpoi2014describing}, Place365\cite{zhou2017places}, LSUN-Crop\cite{yu2015lsun}, LSUN-Resize\cite{yu2015lsun}, and iSUN\cite{xu2015turkergaze}. For few-shot learning, we trained using one-shot and two-shot respectively, and tested on all the OOD datasets.

\paragraph{Setup.} Following the established method\cite{miyai2023locoop}, we utilize the publicly available ViT-B/16 architecture of the CLIP model as our backbone. In the feature map of CLIP, we use a 14x14 patch design. For $\beta$, we fix it at 0.1 to appropriately consider the influence of surrounding information on the patches. We set the number of epochs to 10, learning rate to 0.002, batch size to 8, and the token length in the prompt to 16.

\paragraph{Comparision Methods.}


To validate the effectiveness of our method, we conducted comparisons with zero-shot methods and existing approaches capable of addressing few-shot OOD detection. For the zero-shot methods, we directly utilized MCM and GL-MCM~\cite{ming2022delving} for OOD detection. In terms of the few-shot methods, we compared against CLIPood\cite{shu2023clipood} and LoCoOp\cite{miyai2023locoop}. While CLIPood was originally designed for OOD generalization tasks, it can also be repurposed for few-shot OOD detection tasks. Meanwhile, LoCoOp currently stands out as the best-performing method for few-shot OOD detection. To ensure a fair comparison, we calculated scores using the MCM score\cite{ming2022delving} when determining ID and OOD samples.

\paragraph{Evaluation Metrics.}


To assess the effectiveness of the methods, we compared the performance using the following evaluation metrics: AUROC: the area under the receiver operating characteristic curve.


\subsection{Main Results}




Table~\ref{table:main} reports the experiment results on CIFAR-10 and CIFAR-100.  Our proposed CLIP-OS makes a remarkable performance boost. Table~\ref{table:imagenet} reports the results on ImageNet-100, where CLIP-OS still outperforms the existing few-shot OOD detection methods. From the experiment results, we find several impressive results: \textbf{(1)} As shown in Table~\ref{table:main}, the use of the one-shot method significantly improves OOD detection ability compared to the zero-shot method. This demonstrates the meaningfulness of using few-shot learning to address OOD problems, as it achieves effective improvement with very limited computational resources and data. \textbf{(2)} From the experimental results, it is evident that LoCoOp outperforms CoOp on both CIFAR-10 and CIFAR-100. This underscores the necessity of training with synthetic OOD data. \textbf{(3)} CLIP-OS significantly outperforms existing methods, especially surpassing 78\% on CIFAR-100, while other methods achieve less than 75\%. This proves that our method synthesizes more reliable and effective OOD data by better integrating ID-relevant features. Our ID/OOD boundary regularization is highly effective. \textbf{(4)} Compared to the one-shot method, the improvement from the two-shot method is very limited. This demonstrates the powerful learning capabilities of methods based on large-scale visual language models, indicating that they do not require extensive training data. It also confirms the meaningfulness of few-shot learning.





\begin{table}[t]
\setlength{\abovecaptionskip}{0.1cm}
\centering
\scalebox{1}{
\begin{tabular}{l|c}
\toprule 
ID Data & CIFAR-100 \\
\midrule 
OOD data  &Avg \\
\midrule
\rowcolor{black!20}  CLIP-OS  &\textbf{78.24}\\
w/o Masking   &76.93  \\
w/o Synthesized OOD   &74.12 \\

w/o ``Unknwon'' prompt   &73.40  \\

\bottomrule
\end{tabular}}
\caption{Ablation study on different components of CLIP-OS.}
\label{table:ablation}
\end{table}





\begin{table*}[t]
\setlength{\abovecaptionskip}{0.1cm}
\centering
\scalebox{0.91}{
\begin{tabular}{l|cccccc}
\toprule 
ID Data & \multicolumn{6}{c}{CIFAR-100}\\
\midrule
OOD data   & Textures    & Places   & LSUN-C    & LSUN-R    & iSUN   & Avg  \\
\midrule
Top-K rank   &67.75 &54.27 &89.20  &86.13  &85.59  &76.59 \\
\rowcolor{black!20} Masking (ours)  &\textbf{70.95}  &\textbf{59.04}  &\textbf{89.60}  &\textbf{85.71}  &\textbf{85.88} &\textbf{78.24}\\
\bottomrule
\end{tabular}}
\caption{
Comparison with different methods for ID-relevant features obtaining.
}
\label{table:topk}
\end{table*}

\subsection{Ablation studies}

In this section, we conduct ablation experiments on each component of CLIP-OS to demonstrate its effectiveness.

\begin{figure*}[ht]

\centering
\includegraphics[height=6cm,width=17cm]{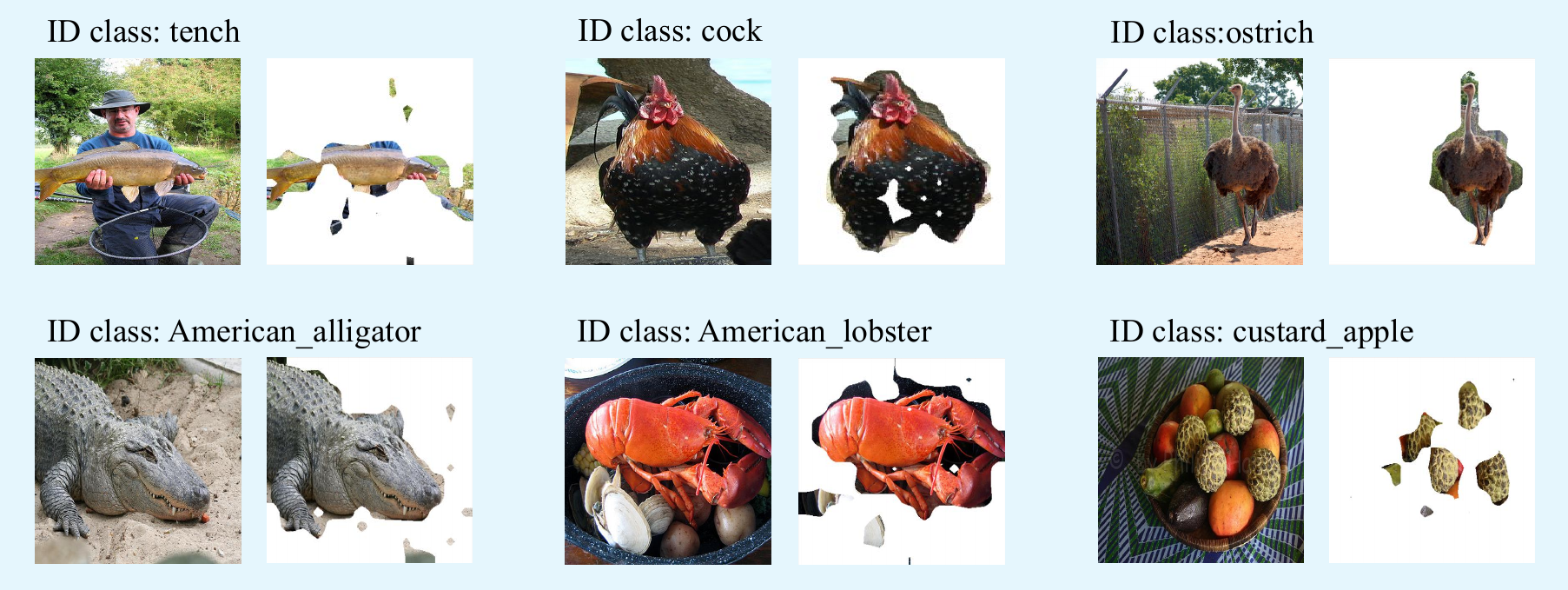}
\caption{Visualization of extracted ID regions.}
\label{fig:visualization}
\end{figure*}

\paragraph{ID-relevant Features Obtaining.} In this part, we evaluate the effectiveness of the ID-relevant Features Obtaining in CLIP-OS and the result is indicated in the second row of Table~\ref{table:ablation}. Without this component, we are only able to perform mix-up on the original images to obtain OOD data, which results in a 3.5\% decline in the average AUROC score. This demonstrates that our approach to obtaining ID-relevant features successfully separates foreground information, thereby mitigating the negative impact of ID-irrelevant data on OOD data synthesis. Figure~\ref{fig:visualization} offers a more intuitive illustration of these findings.

\paragraph{Reliable OOD Data Synthesizing.} We also analyze the impact of the Reliable OOD Data Synthesizing and the result is shown in the third row of Table~\ref{table:ablation}. When OOD data synthesis is omitted, the model lacks essential OOD supervisory signals for training, resulting in a decrease in the average AUROC score from 78.24 to 74.12. This performance gap serves as evidence of the reliability and effectiveness of the synthesized OOD data. The reliable OOD supervisory signals it provides are essential for effective OOD detection training.

\paragraph{``Unknown'' prompt.}

As shown in the third row of Tabel~\ref{table:ablation}, without using the ``unknown'' prompt, we employ entropy maximization as the training loss for OOD samples, leading to a significant decrease in the average AUROC score. This illustrates that our designed regularizing method can mitigate the drawbacks associated with the uniform distribution regularizing. Our method does not impair the performance of ID classification, thereby avoiding a detrimental impact on OOD detection performance.

\begin{figure}[t] 
\setlength{\abovecaptionskip}{0pt}
    \centering
    \includegraphics[width=0.4\textwidth]{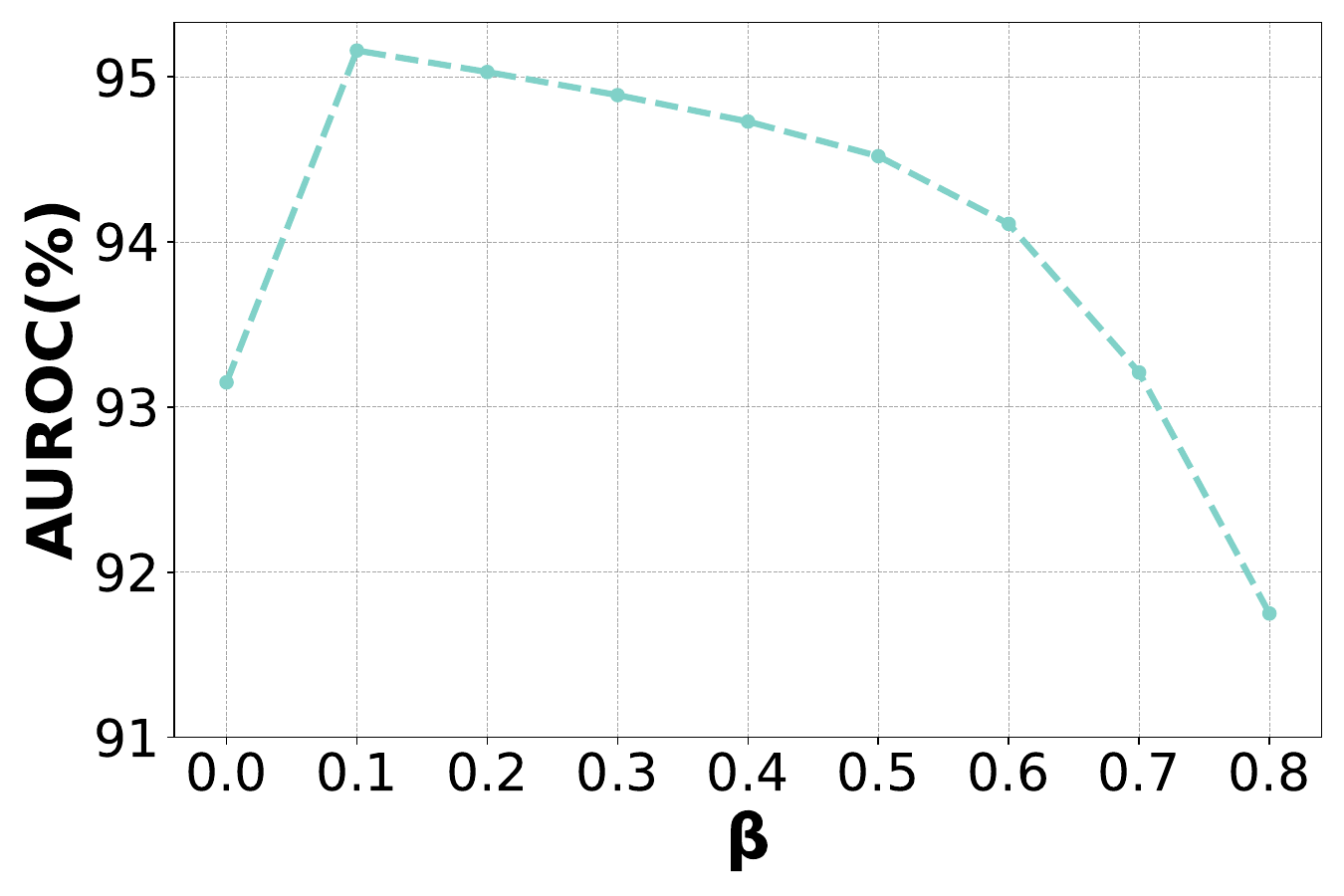} 
    \caption{Sensitivity Experiment of the Parameter $\beta$. We report average AUROC scores on five OOD datasets. }
    

    \label{fig:beta} 
\end{figure}

\subsection{Sensitivity study}

In this section, we conduct a critical sensitivity experiment on a key hyperparameter, $\beta$, which represents the weight of surrounding patches in the patch-context incorporation. We perform experiments using a one-shot setup with $\beta$=0 (Original CLIPsurgery), 0.1...0.8 on CIFAR-10. In Figure~\ref{fig:beta}, we summarize the experimental results, demonstrating the AUROC score under different $\beta$ values. When $\beta$=0, there is a significant decrease in the AUROC score. This indicates that completely disregarding the influence of the surrounding patch information is highly detrimental to obtaining ID-relevant features, thus validating the rationale and effectiveness of our proposed patch-context incorporation. Similarly, when $\beta$=0.5, the AUROC also decreases. This is because excessively considering the surrounding patch information leads to the loss of discriminability among patches' categories, thereby resulting in suboptimal acquisition of ID-relevant features. To balance these considerations, we uniformly use 0.1, which is versatile and does not require adjustment based on different datasets.

\subsection{Further studies}

\begin{figure}[t] 
\setlength{\abovecaptionskip}{0.1cm}
    \centering
    \includegraphics[width=0.46\textwidth]{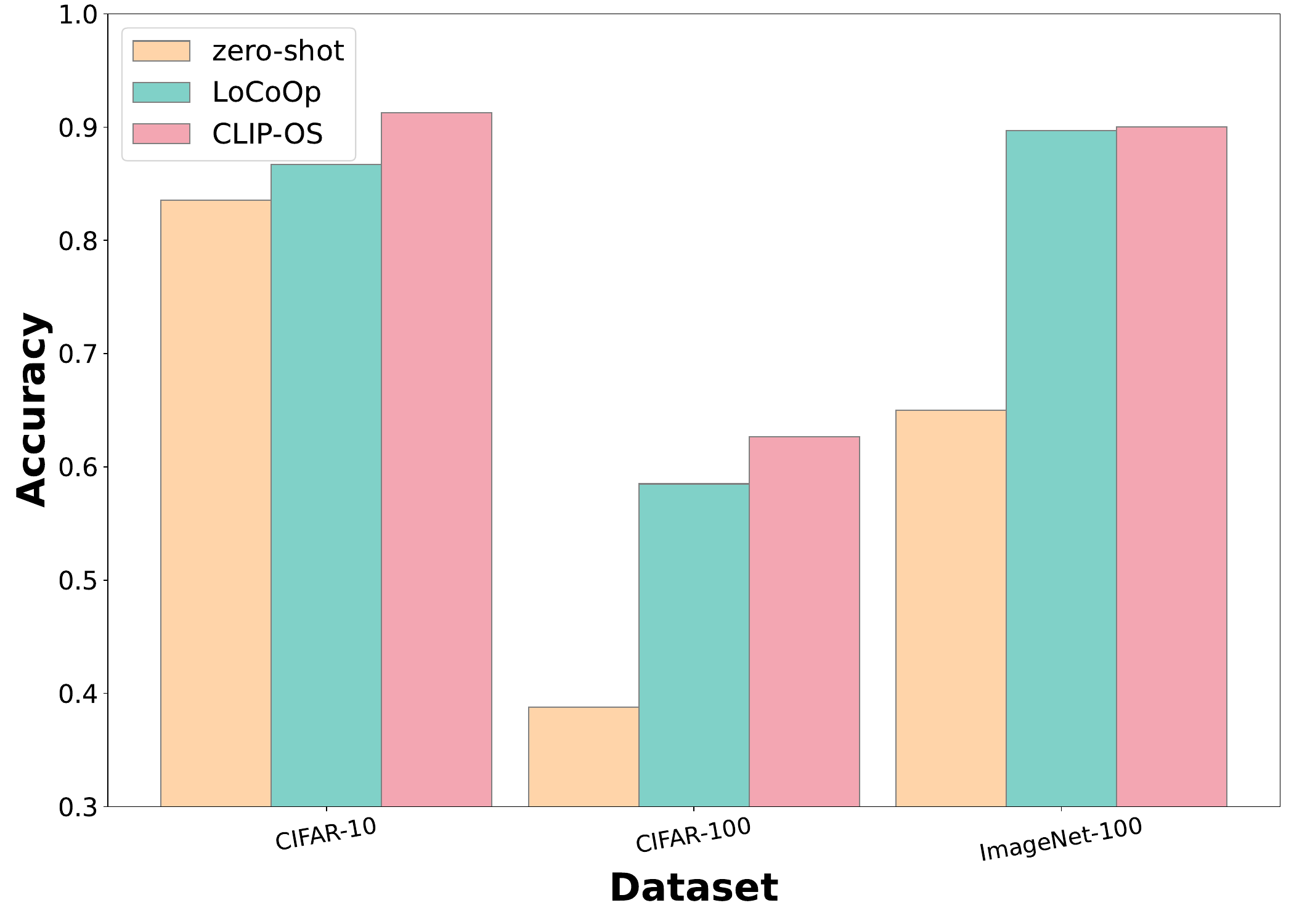} 
    \caption{Comparison in ID accuracy.}
    

    \label{fig:ID_acc} 
\end{figure}

\paragraph{ID classification accuracy of CLIP-OS.}
We evaluate the ID accuracy of CLIP-OS. While our primary focus is on OOD detection, maintaining the model's performance on ID samples is equally important. This is due to the close relationship between ID classification capability and OOD detection ability~\cite{Vaze_Han_Vedaldi_Zisserman_2021}. In Figure \ref{fig:ID_acc}, we illustrate the one-shot ID accuracy of zero-shot, LoCoOp, and CLIP-OS on the test sets of CIFAR-10, CIFAR-100, and ImageNet-100. Our results indicate that LoCoOp exhibits inferior ID accuracy compared to zero-shot, whereas CLIP-OS achieves the highest ID classification accuracy. We attribute this difference to the ID-relevant feature obtaining and unknown prompt introduced in our approach. The former enables the model to train on reliable ID-relevant features and mitigates interference from background information, while the latter, achieved through the use of an ``unknown'' prompt, reduces the impact of OOD detection on ID classification.

\paragraph{Comparison with other methods for ID-irrelevant features obtaining.}

We explore the effectiveness of different methods for ID-irrelevant features obtaining. Despite our implementation of CLIP-surgery-discrepancy Masking, LoCoOp~\cite{miyai2023locoop} offers an alternative method: Top-K rank-based thresholding. Following their methodology, we select patches whose cosine distances to the top $k$ smallest categories text embedding include the ground-truth as foreground. The one-shot OOD detection results are summarized in Table~\ref{table:topk}, where our proposed method achieves superior AUROC scores, indicating that we have extracted more reliable ID-relevant features. Furthermore, it is worth noting that the top-k rank-based thresholding method requires the specification of $k$ for each dataset based on its number of categories, whereas our approach is more generalized.

\paragraph{Visualization of difference mask.} In Figure~\ref{fig:visualization}, we show visualization results of extracted ID-relevant regions. The performance of extraction is key to our method. The results show that our CLIP-surgery-discrepancy Masking can accurately identify ID-relevant regions.

\section{Conclusion}


We presented a novel vision-language-based approach for synthesizing reliable out-of-distribution (OOD) samples in few-shot OOD detection, embodied in our framework named CLIP-OS. Firstly, we extracted ID-relevant features using patch-context incorporating and CLIP-surgery-discrepancy masking. Subsequently, these features are employed to generate reliable OOD virtual samples as OOD supervision signals. Lastly, we proposed an "unknown" prompt to regulate the ID/OOD boundary, minimizing the mutual interference between ID classification and OOD detection. Through extensive experiments, we demonstrated the superior performance of CLIP-OS compared to existing methods.

\newpage

\bibliographystyle{named}
\bibliography{ijcai24}

\end{document}